\documentclass{article}

\usepackage[nonatbib,preprint]{neurips_2024}

\usepackage{graphicx}%
\usepackage{multirow}%
\usepackage[square,sort,comma,numbers]{natbib}
\usepackage{amsmath,amssymb,amsfonts}%
\usepackage{amsthm}%
\usepackage{mathrsfs}%
\usepackage[title]{appendix}%
\usepackage{xcolor}%
\usepackage{textcomp}%
\usepackage{manyfoot}%
\usepackage{booktabs}%
\usepackage{algorithm}%
\usepackage{algorithmicx}%
\usepackage{algpseudocode}%
\usepackage{listings}%

\newcommand\llamatwo[1][]{Llama 2}
\newcommand\gptthree[1][]{GPT-3}
\newcommand\chatgpt[1][]{GPT-3.5}
\newcommand\gptfour[1][]{GPT-4}
\newcommand\bfi[1][]{BFI 2}
\newcommand\openpsy[1][]{50-item IPIP Big Five Markers}
\newcommand\llm[1][]{LLM }

\usepackage[utf8]{inputenc} 
\usepackage[T1]{fontenc}    
\usepackage{hyperref}       
\usepackage{url}            
\usepackage{booktabs}       
\usepackage{amsfonts}       
\usepackage{nicefrac}       
\usepackage{microtype}      
\usepackage{xcolor}         

\title{Challenging the Validity of Personality Tests for Large Language Models}

\author{%
  Tom S\"uhr\\
  Max Planck Institute for Intelligent Systems\\
  T\"ubingen AI Center\\
  \texttt{tom.suehr@tuebingen.mpg.de} \\
   \And
  Florian E. Dorner \\
  Max Planck Institute for Intelligent Systems \\
  ETH Zurich \\
  \texttt{florian.dorner@tuebingen.mpg.de} \\
   \AND
  Samira Samadi \\
  Max Planck Institute for Intelligent Systems \\
  T\"ubingen AI Center \\
  \texttt{samira.samadi@tuebingen.mpg.de} \\
  \And 
  Augustin Kelava \\
  Methods Center \\
  University of T\"ubingen \\
  \texttt{augustin.kelava@uni-tuebingen.de} \\
}

\begin{document}

\maketitle

\begin{abstract}
  With large language models (LLMs) like \gptfour{} appearing to behave increasingly human-like in tex-based interactions, it has become popular to attempt to evaluate personality traits of LLMs using questionnaires originally developed for humans. While reusing measures is a resource-efficient way to evaluate LLMs, careful adaptations are usually required to ensure that assessment results are valid even across human subpopulations. In this work, we provide evidence that LLMs’ responses to personality tests systematically deviate from human responses, implying that the results of these tests cannot be interpreted in the same way. Concretely, reverse-coded items (“I am introverted” vs. “I am extraverted”) are often both answered affirmatively. Furthermore, variation across prompts designed to “steer” LLMs to simulate particular personality types does not follow the clear separation into five independent personality factors from human samples. In light of these results, we believe that it is important to investigate tests’ validity for LLMs before drawing strong conclusions about potentially ill-defined concepts like LLMs’ “personality”.
\end{abstract}

\section{Introduction \& Related Work}
Recent advances in large language models (LLMs) have made these models' responses more and more human-like and have led to an unprecedented amount of human-language-model interactions. This sparked interest in the potential emergence of psychological traits such as psychopathy and personality characteristics like extraversion in LLMs \cite{li2022gpt}

Psychological traits are typically used to describe (more or less stable) habitual human experience and behavior, as well as styles of perception and cognition, and have been studied in humans for decades. 
As part of their development, tests (e.g., for the measurement of cognitive abilities) and questionnaires have been created and discussed by scientists, validated and repeatedly improved. On a meta-level, the process of assessing human traits has been subject to interdisciplinary quality improvement and standardized, for example in the fields of psychometrics and test construction \cite[e.g.,][]{association2014standards}.

As an idea that seems plausible at first glance,
psychological tests are now being used in attempts to assess and quantify (potential) personality traits of LLMs \cite{safdari2023personality, lu2023illuminating, jiang2023evaluating, jiang2023personallm, pellert2023ai,mao2023editing,pan2023llms,huang2023chatgpt,karra2022estimating,singh2023can,huang2023humanity,caron2023manipulating,mei2024turing}. However, like for the performance of machine learning models, it is a priori not clear whether the validity of psychological tests transfers from one population to another (here: humans to LLMs). The fundamental assumption necessary for such a transfer is that the measurement tool (i.e., test or questionnaire) does not change its psychometric properties (i.e., the mathematical functional form between observable behavior and latent/unobserved personality characteristics, as well as its parameters). This assumption is called measurement invariance \cite[e.g.,][]{Meredith1993} and is thoroughly examined \cite[e.g.,][] {danner2016deutsche,gallardo2022factor} when psychological tests and measurement tools are transferred from one human subpopulation to another. This is done to reduce the proliferation of flawed measurement methods and psychological constructs, to which machine learning research is not immune \cite[e.g.][]{adebayo2018sanity}. It also helps to prevent the creation of redundant constructs and trait definitions that are explained through already existing constructs. Negligent transfer of measurement tools to LLMs without such a thorough examination renders certain inferences, such as "consistent responses to a personality test indicate the existence or even specific expression of personality traits (e.g., extraversion)" invalid. Numerous studies employing this practice have been presented at leading machine learning conferences \cite{jiang2023evaluating,huang2023humanity, caron2023manipulating} and major journals \cite{mei2024turing,pellert2023ai}. 
\subsection{Contributions}
In this work, we subject the application of personality questionnaires to LLMs to rigorous tests. We show that LLM responses to the well-known \openpsy{} \cite{ipip50} show patterns that are highly unusual for humans. In addition, we prompt LLMs to imitate a wide range of different ``personas" when responding to a well known questionnaire, namely the \bfi{} \cite{john1991big,soto2017next}.
We find that LLMs fail to replicate the five-factor structure found in samples of human responses. This implies that measurement models that are valid for humans do not fit for LLMs and that currently applied procedures for administering questionnaires to LLMs do not allow for the inference of personality. We hope that this work will guide the analysis and adaptation of other psychometric and educational tests for LLMs.
\subsection{Measurement Invariance and Nomological Nets}
Since the introduction of LLMs, there has been an increasing interest among researchers in investigating the latent characteristics of these models, including personality traits. As a result, the machine learning community has started to incorporate methods from psychometrics to a greater extent. However, quality control measures, analogous to the Standards for Educational and Psychological Testing \cite{association2014standards}, have not been adequately implemented.\\
In this section, our aim is to provide an understanding of why it is crucial to conduct quality control, particularly when examining the validity of psychometric tests for LLMs, before drawing any conclusions. %
For didactic purposes, we will solely focus on two concepts which are particularly important for quality control.
\\
First, is the psychometric function that relates the latent trait (e.g., extraversion) to the multivariate vector of observed behavior (e.g., responses to items) identical across two different subpopulations? For example, are responses of items measuring personality facets in a sample from Boston the same as in a sample from San Francisco? More precisely, are the parameters of the function that generate the answers to the personality test, based on the latent expressions of the personality facets, the same in both samples? If so, we call the personality test \textit{measurement invariant} \cite{Meredith1993}. Without this property, either the underlying model of the latent trait (personality) is incorrect, or the personality test only captures a biased signal of it. 
\\
Second, does the measured latent trait relate to other traits in the new sample as expected? For example, do the results of a new intelligence test correlate sufficiently well with the results of existing intelligence tests? Or if a new latent trait has been introduced, is it really new and, for example, sufficiently independent of other latent traits that already exist? This space of existing constructs and the relationships between them is called the \textit{nomological net} \cite{cronbach1955construct}. Embedding new constructs (models of latent traits) into this space of all constructs prevents the introduction of redundant terms and ensures a certain degree of construct validity, important indicators that a psychometric test measures what it is intended to measure.
\subsubsection{Measurement Invariance} In order to operationalize the investigation of measurement invariance, we first have to introduce some notation.
We are interested in some latent trait $W$ of a population (e.g., a group of humans). $W$ could be something like intelligence, creativity, or personality. $W$ can have more than one dimension (e.g., 5 dimensions in the case of the Big Five personality conceptualization). For now, let $W$ be a $d$-dimensional random variable with realization $w$. We want to find an $n$-dimensional measure $X$ of $W$. In the case of personality, $X$ would be a personality test like the \bfi{} \cite{soto2017next}. We say that $X$ measures $W$. A simple example of a measurement model with our notation would be $X = \lambda W + \varepsilon$. With $\lambda \in \mathbb{R}^{n\times d}$.\\
We call $\lambda$ the factor loading matrix. Every entry $\lambda_{i,j}$ in this matrix corresponds to the covariance between coordinate $i$ of $X$ and coordinate $j$ of $W$. For standardized $W$ and $X$ (with $\sigma^{X}_{ii}=\sigma^{W}_{jj}=1$, the diagonals of the covariance matrices of $X$ and $W$.), the $\lambda_{i,j}$ can be interpreted as correlation. In the case of personality, this would correspond to the signal of the personality facet $j$ (e.g., extraversion) in item $i$ (e.g., "I feel comfortable around people."). We can estimate $\lambda$ using Factor Analysis or PCA or other factor-analytic techniques \cite{rencher10methods}. The error term $\varepsilon$ captures all other factors that influence the values of $X$. On a high level, the characteristics of $\lambda$ and $\varepsilon$ are important for the quality of a psychometric measure. As we will discuss in section ~\ref{pca-cfa-results}, the entries of the factor loading matrix should have high absolute values. We will also discuss why the variance explained by the factor loadings should be sufficiently high compared to the variance explained by the residual noise $\varepsilon$.
However, even if $X$ satisfies these requirements for a sample of the population, it must generalize to all subsamples of the population that have the trait $W$. Otherwise, it could be that $X$ does not measure $W$ but some other trait which is specific to the one sample. 
Let $V$ be the the population of all individuals that have the latent trait $W$.\footnote{Note that we simplified the definition of measurement invariance from \cite{Meredith1993} for didactic reasons.} Let $F(\cdot)$ be the cumulative distribution function and $v\subseteq V$.
We call $X$ measurement invariant if $F(x|w,v)=F(x|w)$ for all $(x,w,v)$. This implies that the factor loading matrix $\lambda$ is independent of the sample on which we estimate it. This property\footnote{Note that weaker forms of measurement invariance exist \cite{Meredith1993} which we leave out for brevity and didactic reasons.} can be checked using so called multi-sample confirmatory factor analysis \cite{brown2015confirmatory}. In other words, for generalizability, it is necessary that a personality test works equally well for every person that \textit{has personality}. Otherwise, we could be measuring something that is specific to the given sample. As we can see, measurement invariance is defined over a population $V$ that has the latent trait of interest. The application of a personality test to an LLM would therefore imply the assumption that the LLM has personality as defined for humans. We will see however, that the retrieved factor loadings on samples of LLM answers to a personality test are substantially different from the factor loadings estimated on human answers which is a violation of minimal requirements. Thus, current LLMs do not necessarily have human personality or the personality tests designed for humans do not measure the same latent variable for LLMs. Both conclusions render the application of personality tests designed for humans to current LLMs futile.
\subsubsection{Nomological Nets}\label{subsubsec:nomological-nets}We call the space of all constructs (models of latent traits) and their relationship to each other nomological nets. Our main focus, when studying nomological nets, is on two types of validity. First, \textit{convergent validity} which refers to the extent to which our measure of interest is related with other measures that it should theoretically be related to. If a new intelligence test is developed, it is expected that the test results from this new test will show a strong correlation with the results obtained from established intelligence tests and other known indicators of intelligence. Second, we want to investigate \textit{divergent validity}, whether a new construct or measure is sufficiently different from already existing constructs and measures. Divergent validity prevents the inflation of new measures and constructs. For example, assume we create a new construct which we call "smartness". Our measurements of smartness correlate strongly with measurements of intelligence and it turns out that smartness predicts job success as good as intelligence. Therefore, the concept of smartness is not significantly different from intelligence, and there is no need to introduce a new term for this latent trait. It is important to note that up until now, nomological nets have been discussed in the context of \textit{human} constructs and measures. Constructing a nomological net for LLMs can lead to a completely different space. Without any further discussion, it is not even clear that the intersection of human and LLM constructs is not empty. This does not mean that LLMs do not posses a latent trait that we \textit{can} describe as "personality". However, technically speaking, we do not know if this construct would have something in common with the construct of human personality. More importantly, measures of the human construct (e.g. personality) can not be used to measure the LLM construct.
\subsection{Population or Individuals}
The objective of psychological assessments is to make inferences about individuals. The validation process of these assessments requires the analysis of groups of people and assessing how effectively the evaluation works within that population.\\
This raises the question, as what we should view LLMs: as individuals or as a populations? Is \gptfour{} a single individual with some variation of answers to the same prompt? Or is the probability distribution over tokens, conditional on a certain prompt, a population from which we simply draw the answer of one individual? Recent works have not come to a conclusion on this matter. Some works treat LLMs as an individual (e.g. \cite{binz2023using}), some as populations (e.g. \cite{santurkar2023whose}). In the context of personality assessments, LLMs have often been treated as a distribution from which they sampled individual personality profiles conditional on predefined \textit{personas}. Instead of simply prompting an item from a personality assessment, they preceded the item with a specification of a persona like "answer as if you are an introvert". Several works employed similar methods to "induce" various personalities into LLMs, for example, by appending "roles" or "personality types". \cite{lu2023illuminating, anonymous2024on, jiang2023personallm, mao2023editing, safdari2023personality} or short descriptions of personalities \cite{jiang2023evaluating}. An advantage of this technique is that it creates variation in the LLM responses and thus enables researchers to analyze them with statistical tools. It is however unclear, what impact this personality induction has on the validity of a personality assessment, even for humans. In this work, we will cover multiple perspectives. First, we will treat current LLMs as individuals to investigate their agree or disagree with a statement independent of it's content (agree bias). Second, we will follow other recent works in the personality assessment of LLMs and view them as populations. We will conduct experiments with and without inducing personas. We release all code and data required to reproduce our experiments (see Appendix \ref{A:data-code}).
\section{Results}\label{sec:experimental_setup}
As a consequence of the explained concepts of measurement invariance and nomological nets, we conducted two experiments in which we tested them.
First, we compared LLM responses to a large sample ($n=1,015,342$) of human online responses\footnote{Data of human resonses can be found at \url{https://openpsychometrics.org/tests/IPIP-BFFM/}} on the \openpsy{} \citep{ipip50}. In this experiment, we treat LLMs as individuals without induced personas. We found that LLMs respond inconsistently to questions that aim to measure the same latent personality facet (e.g. extraversion). More specifically, they both agree to opposite items like "are you an extrovert" as well as "are you an introvert".
Second, we recorded responses to the \bfi{} \citep{soto2017next}, an updated version of the BFI, for LLMs prompted to imitate different personas. We used the personas from \cite{safdari2023personality}, for more details please consult ~\ref{si:personas}. Additionally, we conducted the experiment without using personas. The experiments were carried out in two configurations: "no-context," where the LLMs were queried with each item in a new context window, and "in-context," where the LLMs were queried with each item in a context window containing all previous items and the LLM's responses to those items. For a detailed explanation of the prompt setup, please consult \ref{si:prompt}. We examine the data of these 2x2 (personas,no-personas) x (no-context, in-context) treatments with standard tools for quality control in psychological testing, exploratory and confirmatory factor analysis as well as reliability measures. Our results show that the \bfi{} is not measurement invariant between humans and LLMs, rendering it's interpretation as a measure of personality invalid.
\begin{figure*}[t]
\begin{center}
\includegraphics[width=\linewidth]{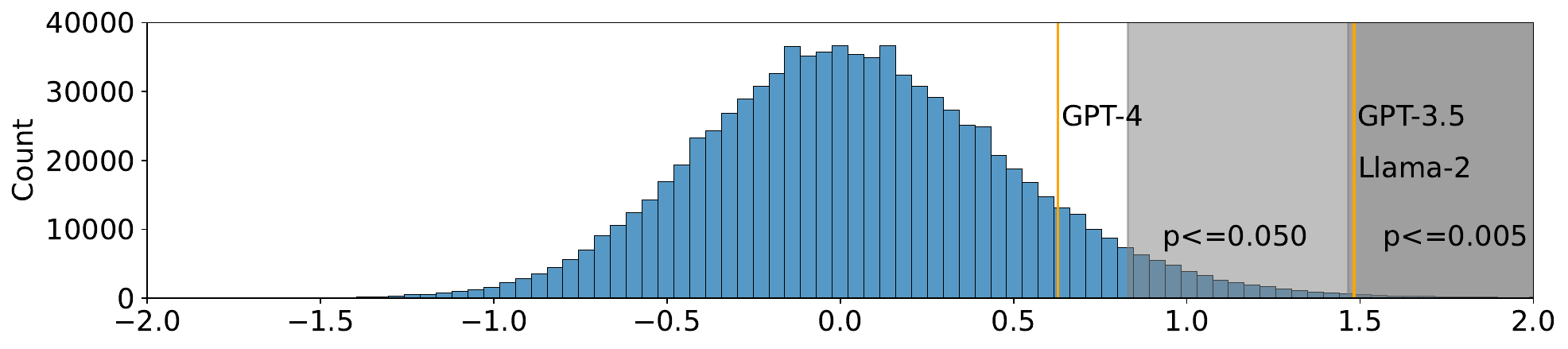}
\end{center}
\caption{Histogram of Agree Bias $a_i$ in human sample, compared to LLMs.}\label{fig:hist}
\end{figure*}


\subsection{LLMs show unhumanly agree bias}\label{sec:results-bias}
For our first experiment on the \openpsy{}, we focus on what we call \textit{Agree Bias}, the tendency of LLMs to produce answers that signify agreement independent of the actual item. To assess this bias, we first convert the scores $s(x)$ for both \textit{true key} (for example assessing extraversion) and \textit{false key} items (for example assessing introversion) $x$ to a single common scale (for example measuring extraversion) by ``flipping" the scores for false key items, setting:
\begin{equation}s^{c}(x)=\begin{cases}
s(x) & x \text{ true key} \\
6-s(x) & x \text{ false key}
\end{cases}.\end{equation}
By design, we expect $s^{c}(x)$ to be similar for true- and false key items for human respondents, while a simple bot that always answers with ``Agree" would have $s^{c}(x)=5$ for true key and $s^{c}(x)=1$ for false key items. Correspondingly, we define a respondent $i$'s agree bias as the average score $s^{c}_i(x)$ for true key items minus the average score for false key items:
\begin{equation}  a_i = \sum_{x \in \text{ True key}} \frac{s^{c}_i(x)}{|\text{True key}|} -  \sum_{x \in \text{ False key}} \frac{s^{c}_i(x)}{|\text{False key}|}.\end{equation} 
Figure \ref{fig:hist} shows the histogram of agree biases in the human sample, as well as the agree biases for the prompted LLMs. As expected, the average agree bias for humans is close to zero. Meanwhile, all LLMs exhibit clear agree bias ranging from $0.6$ for \gptfour{} to $1.5$ for \llamatwo{} and \chatgpt{}. For the latter two, we can reject the null hypothesis ``the model's agree bias is sampled from the same distribution as human's agree biases" at $p<0.005$, using the human sampling distribution for a model-free hypothesis test. While the results for \gptfour{} are not statistically significant, they remain suggestive with the model's agree bias exceeding 89\% of humans' agree biases.
\subsection{The BFI 2 does not capture the five dimensions of human personality in LLMs}
For our second experiment, we test each LLM on the \bfi{} for each of the following settings: i) no-context persona prompts, providing a persona (e.g. "I love to hike.")  with every query and for each item in a new context window.
ii) in-context persona prompts, in which we ask all items of the \bfi{} in one context window for each persona. iii) in-context seeded prompts. In this setting, we set the answer to the first question to one of the five options and then query the LLMs with the remaining 59 questions of the \bfi{} in one context window. In this setting, personas were not added to the prompt. We collected $n=100$ completed questionnaires for each setting.
We also tested in-context and no-context settings without personas or seeding. However, missing variance (even with higher temperature settings) prevents an exploratory and confirmatory factor analysis. More details about the prompt setup can be found in \ref{si:prompt}.\\
PCA (as well as exploratory factor analysis) is a useful tool to reduce the dimensionality of the observed item responses to components\footnote{Note that in exploratory factor analysis, these components are conceptualized as real latent/unobserved variables.}. The dimensionality reduction results in factors that can be interpreted as underlying dimensions (here: ideally personality factors). As an exploratory approach, an estimate of a full factor loading matrix is obtained (based on a solution of eigenvalues) that describes the relationship between observed items and components. If the items do not show a so-called simple structure, i.e., high loadings only on the factors they are supposed to measure and zero loadings on the other factors, then items are considered to be heterogeneous.
For each of the LLMs and prompt setups, we conduct a PCA with varimax rotation of the standardized item scores $s(x)^{std}$ for item $i$ to obtain the model%
\begin{equation}\label{pca-cfa-results}
s(x)^{std} \approx \sum_{g=1}^5 \lambda'_{gx} \xi_{g} 
\end{equation} where $\lambda'_{gx}$ is called the \textit{component loading} of item $x$ for the component $g$, while $\xi_{g}$ represents the value of component $g$ for a given persona/individual\footnote{Note that we omit the index that represents the persona/individual here.}. By design of the \bfi{}, the PCA has two important characteristics on human data: First, as each of the Big Five factors describes a clean and somewhat orthogonal axis of variation in human behavior, we expect each of the five learnt components $g$ to have a strong association with items from exactly one of the Big Five factors, yielding a simple structure for $\lambda'$. Indeed, the \bfi{} was in part designed to fulfill this property, which can be achieved by removing items that correlate with multiple components during test design. Second, as affirmative answers to false key items are supposed to indicate adherence to the opposite end of a component-spectrum as true key items, we expect the component loadings $\lambda'_{gx}$ for false key items $x$ that belong to component $g$ to have the opposite sign as the corresponding true key items. 
\begin{figure*}[t]
    \centering
    \includegraphics[width=\linewidth]{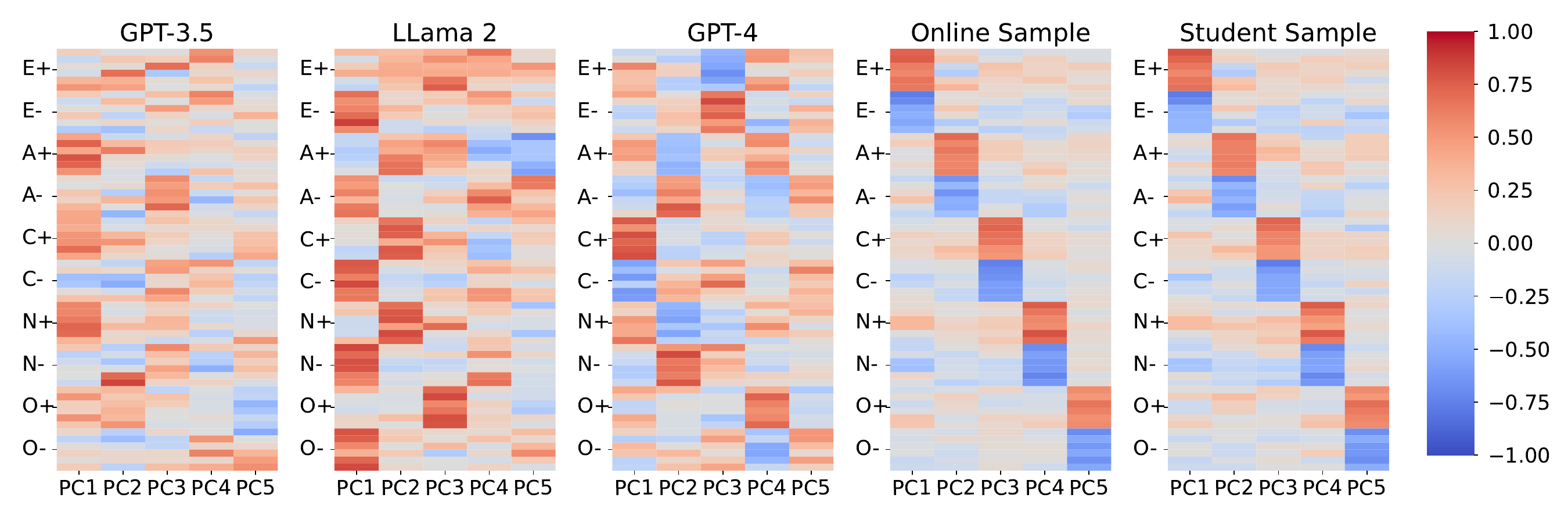}
    \centering
    \caption{Component loadings of PCA with Varimax rotation for LLM (no-context prompting with personas) and human samples of \cite{soto2017next}. +, - indicate true- and false-key items of the \bfi , letters stand for Big Five factors.}
    \label{fig:pca}
\end{figure*}%
\begin{figure}[ht]
    \centering
    \includegraphics[width=\linewidth]{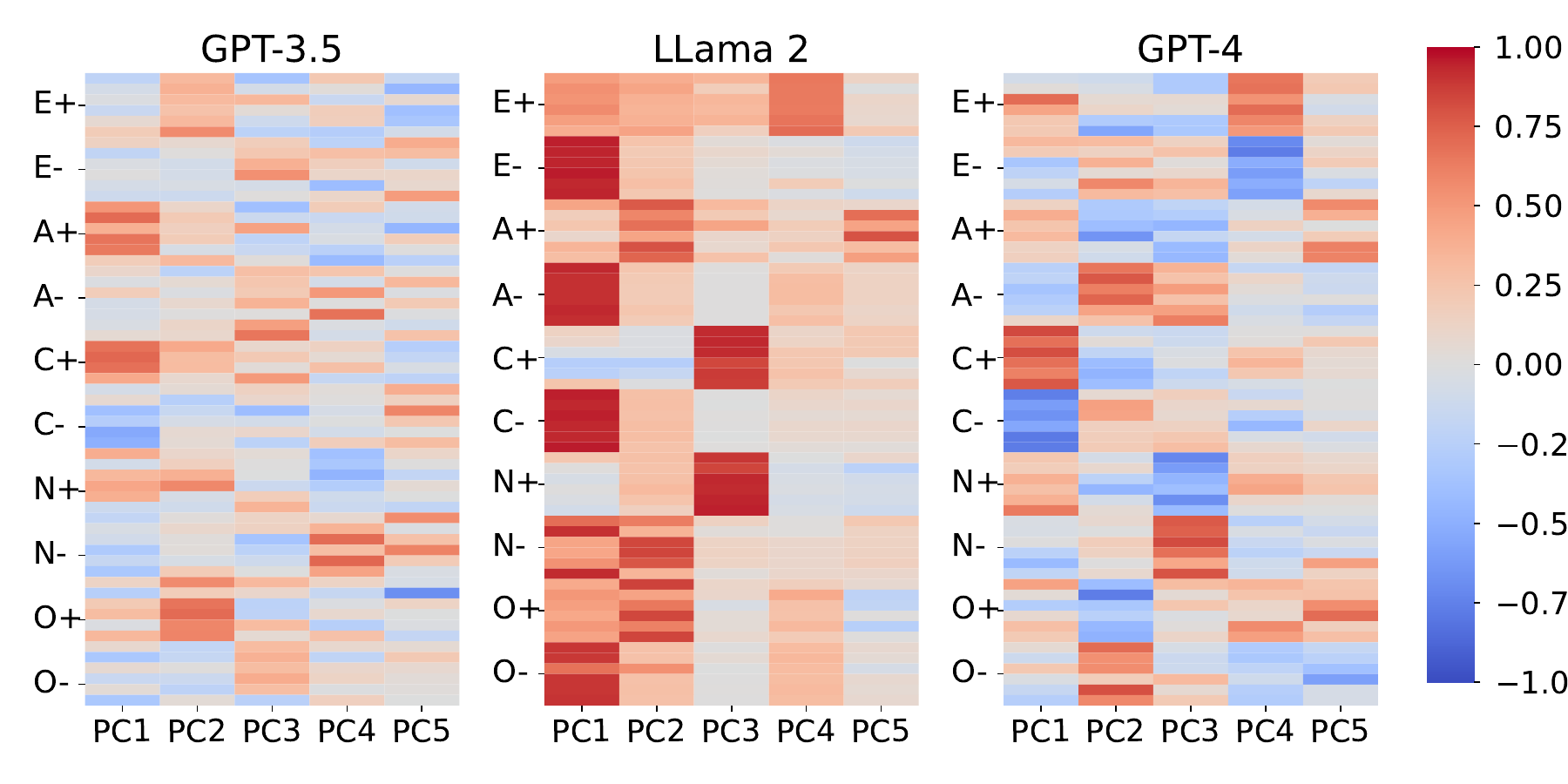}
    \centering
    \caption{Component loadings of PCA with Varimax rotation for LLMs of incontext prompting with personas. +, - indicate true- and false-key items of the \bfi , letters stand for Big Five factors.}
    \label{fig:pca-incontext}
\end{figure}%
\begin{figure}[ht]
    \centering
    \includegraphics[width=\linewidth]{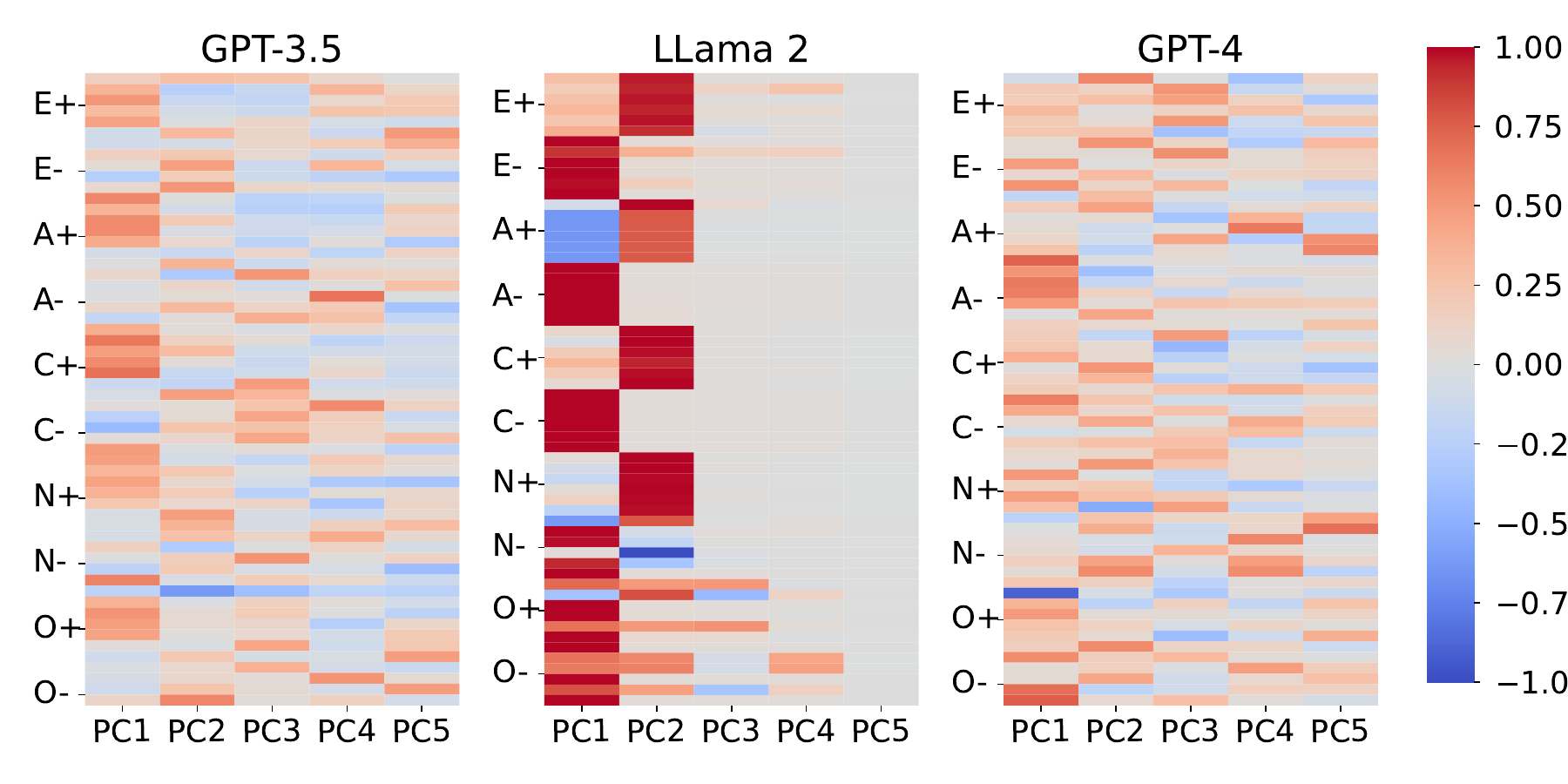}
    \centering
    \caption{Component loadings of PCA with Varimax rotation for LLMs of incontext prompting with first answer seeded. +, - indicate true- and false-key items of the \bfi , letters stand for Big Five factors.}
    \label{fig:pca-incontext-seeded}
\end{figure}%
Figure \ref{fig:pca} shows the component loadings for the LLMs in the no-context setting with personas, compared to the corresponding loadings obtained for human populations using the same procedure \citep{soto2017next}. We only find limited true vs. false key separation for \gptfour{} and not the other two models. Figure \ref{fig:pca-incontext} shows the component loadings of the LLMs in the in-context setting with personas, showing also limited signs of a true-false key separation for \gptfour{} and higher loadings for \llamatwo{} compared to the no-context persona prompting. Figure \ref{fig:pca-incontext-seeded} depicts the component loadings without personas and the seeded answer, showing even less true-false key separation for \gptfour{} and a 2-component structure for \llamatwo{}. 
Most crucially, \textit{none} of the models exhibits the clean block structure (simple structure) in any of the prompt settings, which was intended in the design of the \bfi{} and found in the human samples (seen in Figure \ref{fig:pca}).
This structural deviation between humans and LLMs implies that test validity does not transfer or in other words, the \bfi{} is not measurement invariant between humans and LLMs.\\ Nonetheless, as this conclusion is solely based on the observations presented in Figure \ref{fig:pca}, it is essential to consider additional quanitfiable assessments. Therefore, we aim to proceed with our investigation by performing a confirmatory factor analysis to obtain a measurable outcome regarding the patterns observed in the factor loading matrices.
\vspace{-2mm}
\subsubsection{Confirmatory Factor Analysis}\label{subsec:cfa} A confirmatory factor analysis (CFA) allows for the estimation of a sparse loading matrix, where loadings are fixed to zero a priori and not estimated. CFA is typically used to examine if a measurement model holds in different populations (using model difference tests). For the sake of brevity (and because the models do not even have a good model fit for themselves), we will only discuss the most important fit indices, Comparative Fit Index (CFI), Tucker-Lewis Index (TLI) and Root Mean Square Error of Approximation (RMSEA). CFI and TLI scores of $\geq 0.95$ and RMSEA scores of $\leq 0.06$ are considered acceptable \cite{hu1999}.
\begin{table}[t]
\begin{center}
\scalebox{0.61}{
\centering
\begin{tabular}{clllllllllll}
\multicolumn{1}{l}{} & \multicolumn{2}{c}{} & \multicolumn{3}{c}{\textbf{Single Component}} & \multicolumn{3}{c}{\textbf{3 Sub-Components}} & \multicolumn{3}{c}{\textbf{Full Model}} \\ \hline
\textbf{Model} & \multicolumn{1}{c}{$\alpha$} & \multicolumn{1}{c}{$\omega_h$} & \multicolumn{1}{c}{CFI} & \multicolumn{1}{c}{TLI} & \multicolumn{1}{c}{RMSEA} & \multicolumn{1}{c}{CFI} & \multicolumn{1}{c}{TLI} & \multicolumn{1}{c}{RMSEA} & \multicolumn{1}{c}{CFI} & \multicolumn{1}{c}{TLI} & \multicolumn{1}{c}{RMSEA} \\ \hline
\llamatwo & \multicolumn{1}{l}{\textbf{0.94} / \textbf{0.85}} & \textbf{0.96} / \textbf{0.87} & 0.49 / 0.47 & 0.38 / 0.35 & 0.65 / 0.33 & 0.61 / 0.46 & 0.52 / 0.34 & 0.58 / 0.33 & 0.32 / 0.29 & 0.30 / 0.26 & 0.39 / 0.20 \\ \hline
\chatgpt & \multicolumn{1}{l}{\textbf{0.72} / \textbf{0.70}} & \textbf{0.73} / \textbf{0.70} & 0.74 / 0.51 & 0.68 / 0.39 & 0.10 / 0.17 & 0.39 / 0.39 & 0.25 / 0.25 & 0.15 / 0.18 & 0.36 / 0.24 & 0.33 / 0.21 & 0.10 / 0.13 \\ \hline
\gptfour & \multicolumn{1}{l}{\textbf{0.92} / \textbf{0.90}} &\textbf{0.92} / \textbf{0.90} & 0.78 / 0.74 & 0.73 / 0.68 & 0.13 / 0.18 & 0.55 / 0.50 & 0.45 / 0.39 & 0.25 / 0.25 & 0.58 / 0.53 & 0.56 / 0.51 & 0.12 / 0.12 \\ \hline
Human & \multicolumn{1}{l}{\textbf{0.87}} & \multicolumn{1}{c}{\textbf{0.79}**} & 0.79 & 0.74 & 0.13 & 0.90 & 0.87 & 0.09 & \multicolumn{1}{c}{0.71*} & \multicolumn{1}{c}{0.70*} & \multicolumn{1}{l}{0.07*}
\end{tabular}}
\caption{Reliability scores ($\alpha$, $\omega_h$) and fit indices (CFI, TLI, RMSEA) per LLM (with personas). Values on the left side of the forward slash are in-context results, on the right side are no-context results. Values are averaged over all personality traits for the Single Component and 3 Component model. Acceptable scores are bolded. Human data from \cite{soto2017next}. *CFA on human responses to IPIP Big Five Markers to establish a human baseline.**Value from human data from the german version of the \bfi{} \citep{danner2016deutsche}. Note that all reliability scores are not meaningful due to poor model fit, even though the scores are in an acceptable range.}\label{tbl:reliability-scores-short}
\end{center}
\end{table}
For the data of our second experiment, we conducted a CFA\footnote{We use the lavaan \url{https://lavaan.ugent.be/} package in R for all our CFA} for four models of personality per LLM and prompt setting. First, for a "single component" model of personality which assumes that all items of one personality trait load on a single factor. Second, a "three sub-component" model which assumes that the items of each trait load on three sub scales. For example, the three sub-scales of Extraversion are Sociability, Assertiveness and Energy Level \cite{soto2017next}. Table \ref{tbl:reliability-scores-short} shows the mean fit parameters for both models for the prompt setups with personas in-context and no-context. For all LLMs, the responses to the \bfi~have poor fit to the single component factor model. This reflects the component-wise results of our exploratory factor analysis in Figure ~\ref{fig:pca}, where \chatgpt{} shows no signs of a simple structure and \gptfour{} shows some item-specific block structures. \gptfour{} reaches the best fit indices among LLMs, comparable to humans. However, both human and \gptfour{} fit indices are far below acceptable levels. Extending the model from a single factor to three sub-factors, improves the fit of \llamatwo{} responses in-context, but worsens it for all other LLMs and context settings. The fit of human responses improves close to acceptable levels. In the original paper of the \bfi{} \cite{soto2017next}, Soto and John extend the 3 sub-component model by another "acquiescence" factor. This factor accounts for the tendency to agree/disagree with items. With this additional factor, the \bfi{} reaches acceptable or very close to acceptable levels for both human samples\footnote{The sub-optimal fit scores of the BFI are one reason why it is subject to discussions.}. However, no LLM prompt setup produced responses that were able to converge for this "3+1" model, rendering the calculation of fit indices impossible. 
Finally, we conduct a CFA with five factors and all items of the \bfi. This model aims to quantify the PCA seen in Figure \ref{fig:pca} and Figure \ref{fig:pca-incontext} as a whole. Accordingly, all LLM responses to the \bfi~have poor fit. The CFA of the in-context prompts without personas can be found in Table \ref{si-tbl:seeded-cfa}. Only the single component models were able to converge to a solution. Here too, with non-satisfactory fit indices. These result, as well as the other (attempted) CFAs, confirm the visual result of the PCA, that the \bfi{} is not measurement invariant between humans and the tested LLMs. For the details of the CFA, we refer to Appendix~\ref{si:details-cfa}.
\subsubsection{Steps of Invariance} In the validation process of a psychometric, satisfactory fit indices are just one of multiple requirements to certify measurement invariance. The CFA tells us how well the data can be explained by our hypothesized model. However, the parameters of our model could differ between groups (or between LLMs). Thus, we would have to conduct a "multi-group CFA" in which we restrict the parameters of the model to be equal for all groups and test it against a model where a subset of parameters are allowed to be unequal across groups. In this analysis, we can restrict only the factor loadings (metric invariance), factor loadings and intercepts (scalar invariance) or factor loadings, intercepts and residual variances (invariant uniqueness). These differentiations of invariance can help to account for group specific factors. However, in order to compare parameters between groups, it is necessary that the CFA within each group has at least satisfactory fit indices (otherwise there is no need to test it against an even sparser model with equal parameters across groups which leads to an even worse model fit and less mearsurement invariance). Future work could investigate the differences of latent traits between groups of LLMs (e.g. LLama models vs GPT models) or between humans and LLMs. However, because no LLM data has acceptable fit for any model of human personality, we can not conduct a multi-group CFA in this work.
\subsection{The interpretation of reliability measures is invalid}\label{subsec:reliability} To compare with previous work on personality tests for LLMs \citep{safdari2023personality}, we attempted to estimate the reliability of the \bfi{} using two standard scalar measures, Cronbach’s $\alpha$ and McDonald’s hierarchical $\omega_h$ \citep{mcdonald1999test,zinbarg2005cronbach}.
%
%
%
%
The sum score $S$ of a given scale (e.g., Extraversion) is defined as 
\begin{equation}
    S= (\sum_{i=1}^k \lambda_{i}) \cdot \eta + \sum_{j=1}^3 (\sum_{i=1}^{k(j)} \lambda_{ij}) \cdot \xi_j + \sum_{j=1}^3 \sum_{i=1}^{k(j)} \varepsilon_{ij}
\end{equation}
Generally speaking, reliability is the proportion of variance in the sum score (scale) that can be explained by the (general) factor we intend to measure. Accordingly, $\omega_h$ is defined as:
\begin{equation}\label{eq:omega_h}
    \omega_h = \frac{(\sum_{i=1}^k \lambda_{i})^2 \cdot \text{Var}(\eta)}{\text{Var}(S)}
\end{equation}
with
\begin{equation}
    \text{Var}(S) = (\sum_{i=1}^k \lambda_{i})^2 \cdot \text{Var}(\eta) + \sum_{j=1}^3 (\sum_{i=1}^{k(j)} \lambda_{ij})^2 \cdot \text{Var}(\xi_j) + \sum_{j=1}^3 \sum_{i=1}^{k(j)} \text{Var}(\varepsilon_{ij})
\end{equation}
Following \cite{zinbarg2005cronbach}, Cronbach's $\alpha$ is a special case, where we assume $\lambda_{1} = \cdots =\lambda_{k}$ and $Var(\xi_j)=0$ for all $j=1...3$. However, the interpretation of $\omega_h$ as a measure of reliability relies on the assumption that a hypothesized hierarchical structural (equation) model (i.e. three sub-components for each Big Five factor) accurately represents the data. This assumption can be tested using Confirmatory Factor Analysis \citep[CFA; e.g.,][]{bollen1989structural}. As interpreting $\alpha$ as a measure of reliability relies on even more stringent assumptions than for $\omega_h$, neither $\alpha$ nor $\omega_h$ are meaningful if the CFA fails. Correspondingly, we conducted a CFA for the data from each of the LLMs for each facet of the \bfi~data.\\
Table \ref{tbl:reliability-scores-short} shows the mean reliability scores computed on the single component structural equation models. 
Worryingly, the calculated $\alpha$ and $\omega_h$ are all acceptable ($\geq .7$) and quite large for \llamatwo{} and \gptfour{}, which would be easy to mistake for a sign of good reliability. However, as discussed in the previous section, the CFA revealed an unacceptable fit of the structural model for either of the LLMs on any of the five main facets.
This demonstrates that scalar reliability indices should not be taken at face value when the fundamental assumption of an adequate fit of the underlying structural model is not established. It underscores the necessity of prioritizing model fit assessment, and thus construct validity, before drawing conclusions from values of $\alpha$ or $\omega_h$ on their own.
%
%
\section{Limitations \& Broader Impact}
This section discusses limitations and broader impacts. Although primarily critiquing interpretation of LLM responses to personality tests, simulating human responses with LLMs may be promising for item discovery \cite{petrov2024limited}. Furthermore, although we argue that LLMs likely lack human personality traits, future research could explore and define new "LLM personality" constructs using our methods. Additionally, we test a limited number of models due to resource and space constraints, and results may vary across different LLMs. Our current understanding and scope of this work may not fully capture the broader implications. We recognize the potential misuse of fine-tuning LLMs to mimic human behavior in various contexts. Nevertheless, we believe the benefits of addressing a significant methodological flaw in the evaluation and benchmarking of LLMs outweigh the possible negative impacts.
\section{Conclusion}\label{sec:discussion}
In this work, we have provided evidence that personality tests do not generalize to LLMs.
We found agree bias among LLMs on the \openpsy{} test that would be unusally high for humans. Prompted with and without the instruction to simulate a range of personas,  LLMs failed to replicate the clean structure of variation found in human responses on the \bfi{}. A confirmatory factor analysis quantitatively confirmed these results.

The agree bias could be an artifact of the Reinforcement Learning From Human Feedback (RLHF) \citep{ouyang2022training} employed for training all of the models we considered, and a tendency of human annotators to prefer models that agree with them. However, it also points towards deeper issues with interpreting answers of LLMs to psychological tests: If our measure of a model's ``extraversion" already depends strongly on whether we use true- or false key items in a survey, it appears unlikely that LLMs' ``extraversion" can be extrapolated beyond specific personality surveys.

Like \cite{safdari2023personality}, we find acceptable values of scalar measures of reliability such as $\alpha$ and $\omega_h$. However, in all prompt settings and for all LLMs, they are considerably lower compared to PaLM models on the IPIP-Neo-300.  Meanwhile, a confirmatory factor analysis (CFA) suggests that the factor model on which the calculation of $\omega_h$ is based does not provide adequate fit on our LLM data, such that $\omega_h$ and $\alpha$ cannot be interpreted as a measure of reliability. This discrepancy in results could be due to one of two reasons: a) 
The IPIP-Neo-300 could yield better fit of the factor model $\omega_h$ is based on for LLMs. It could also be more reliable than the \bfi{}, for example because of the large number (300) of test items on the IPIP-Neo-300 and the general tendency of reliability to increase with increasing test length \cite{spearman1961proof,wainer2001true} or b) PaLM could be better than the models we considered at simulating distributions of human personality and thus yield sufficient fit for the factor model underlying $\omega_h$ as well as better scores.).
Together, our results suggest that validity has to be examined critically before a psychological test is applied to a LLM, as validity does not appear to hold for at least one combination of psychological test and state of the art LLM. Validity cannot be assumed when applying psychological tests to new language models without a thorough and critical analysis. Taking a step back, our results provide evidence that while tests designed for humans provide a cheap way of evaluating language models, the results of these evaluations can be misleading, as the tests are built to differentiate \textit{humans from other humans, not language models from other language models or humans}. Such misleading assessments can be problematic as they may obscure important issues with LLMs that do not get caught by the assessment while simultaneously diverting resources and attention towards false concerns arising from flawed tests. 
For example, flawed assessments could erroneously identify alarming traits such as psychopathic tendencies in LLMs and trigger costly mitigation measures. At the same time, LLMs performance on certain tasks might be strongly overestimated based on some LLMs' strong results in academic test, leading to costly mistakes due to premature deployment. 
\textit{If} language models behaved sufficiently human-like in a particular domain, human tests could still provide a lot of useful information, but similarities to humans would have to been established on a case by case basis, and can in particular not usually be concluded \textit{based on the tests' results themselves}.
\bibliographystyle{unsrt}
\bibliography{sn-bibliography}
\appendix
\section*{Appendix}
\section{Data \& Code}\label{A:data-code}
All processed LLM responses and our code for prompting and analysis have been uploaded to Github at \textbf{obfuscated for double blind peer review. code and data has bin submitted as .zip file}
\section{LLM and Hardware Details}\label{si:llm-hardware}
We query the June version of GPT 3.5 (gpt-3.5-turbo-0613) and GPT 4 (gpt-4-0613) and the LLama-2-70b-chat version published at \url{https://huggingface.co/meta-llama/Llama-2-70b-chat}. For LLaMA2, we use the huggingface implementation\footnote{\url{https://huggingface.co/docs/transformers/model_doc/llama2}}, querying the model in 32-bit using four 80-GB A100 GPUs. With this hardware, our LLaMA2 experiments required $\sim$24hours and the GPT experiments 2-3 hours. 
\section{Prompt Engineering}\label{si:prompt}
\begin{figure}[ht]
\includegraphics[width=\linewidth]{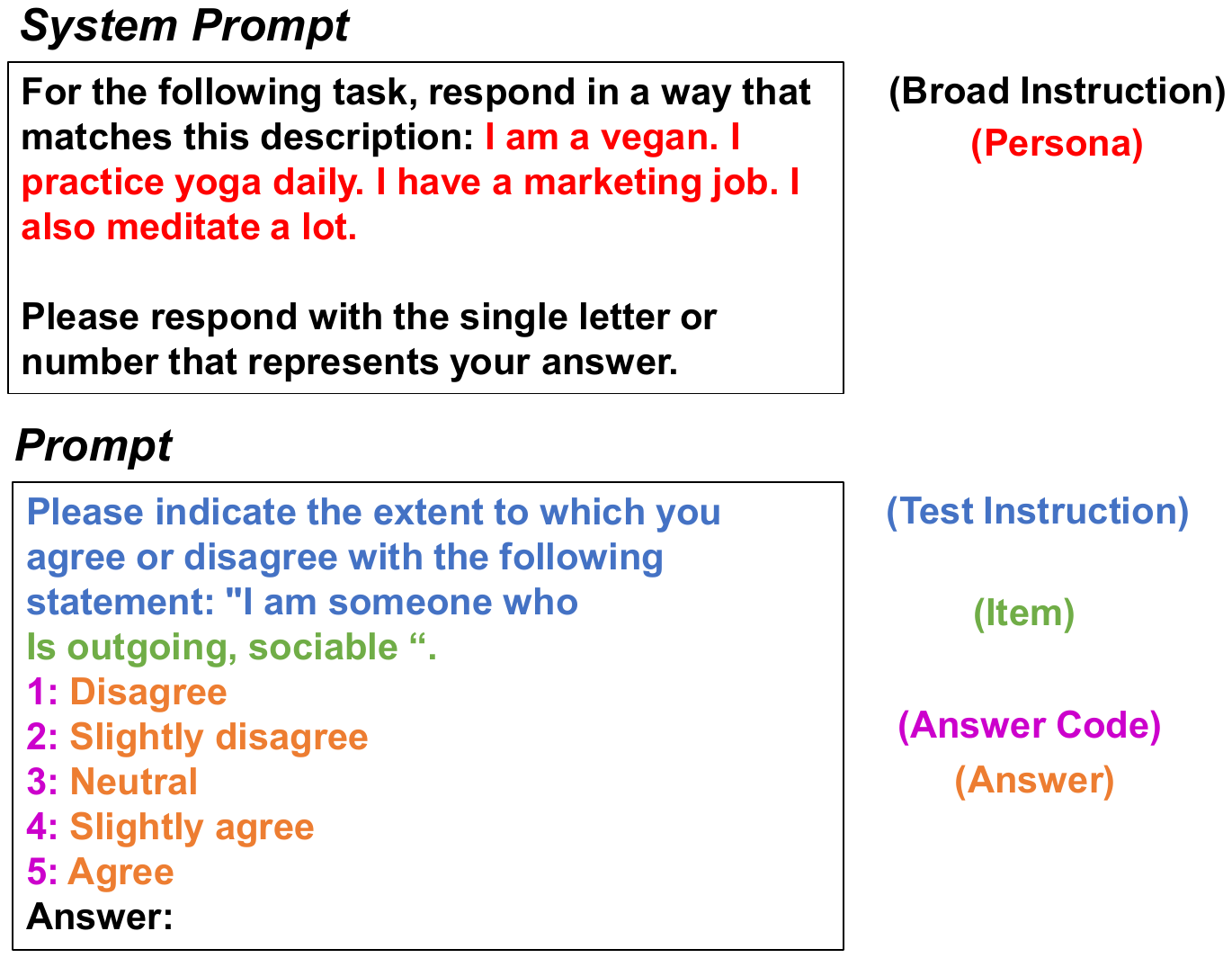}
\caption{Prompt Template}\label{fig:prompt-example}
\end{figure}

Figure \ref{fig:prompt-example} shows the building blocks of our prompts. The system prompt contains one of 100 personas (which consist of a few short statements about a fictional person) as well as the request to answer with a single letter or number. The system prompt in our experiments with empty personas only contained the request to answer with a single letter or number. The main prompt contains the test instruction from the \bfi{} \cite{soto2017next} (\textit{``Please indicate the extent to which you agree or disagree with the following statement: "I am someone who " ``}.) The test instruction is directly followed by the item. After the items, we list the answer codes and the corresponding description of that code (answer) from the \bfi{}. The prompt ends with the ending \textit{Answer:}. For the "in-context" experiments, we append the previous questions with the previous answers before the new item.\\
For \chatgpt{} and \gptfour{} we used the OpenAI chat API. We set temperature to zero, max\_tokens to one and 
\begin{verbatim}
messages = [{"role": "system", "content":
system_instruction},
{"role": "user","content": survey_item}]
\end{verbatim} 
where system\_instruction represents the system prompt and survey\_item represents the prompt. We then record the answered token if it matches one of the answer code tokens, and a non-response otherwise. In our analysis, we map non-responses to the score $s(x)=3$.\\
\\
For LLaMA2, we use the template provided in \url{https://huggingface.co/blog/llama2} to separate the system instructions from the prompt:  
\begin{verbatim}
<s>[INST] <<SYS>>
{{ system_instruction }}
<</SYS>>

{{ survey_item }} [/INST]
\end{verbatim} where again system\_instruction represents the system prompt and survey\_item represents the prompt. We predict the next token based on this input and apply a softmax to the corresponding logits $l$ to obtain probabilities $p'$. We then collect the subset of tokens $\{p'_t, t\in T\}$ that corresponds to the answer code tokens $t$ and renormalize to obtain $p_t = \frac{p'_t}{\sum_{j\in T} p'_j}$. We attempted to conduct our experiments without personas and without seeding the first answer. For \chatgpt{} and \gptfour{} we tried temperatures 0, 0.5 and 1. All three settings yielded no variance in most of the item responses.
\section{Personas}\label{si:personas}
As \cite{safdari2023personality}, we use 100 personas from the PersonaChat Dataset \cite{zhang2018personalizing}. "Empty persona prompts" simply have no persona in the system prompt and no instruction to respond according to a persona. The full list of personas can be found in our github repository \textbf{obfuscated for double blind peer review. code and data has bin submitted as .zip file}
\section{Details PCA}\label{si:details-pca}
For the PCA, we used the \textit{psych} package in R\footnote{https://personality-project.org/r/psych/help/principal.html}. We applied varimax rotations in all PCA's. Varimax is an orthogonal rotation which was also used in the analysis of the original \bfi{} paper \cite{soto2017next}. In summary, the PCA command in R is
\begin{verbatim}
    ncomp <- 5
    pca_rotated <- psych::principal(response_data,
    rotate="varimax", nfactors=ncomp, scores=TRUE)
\end{verbatim} 
with ncomp defining the number of components we are looking for. In the case of personality we set this value to five in order to retrieve the five facets of personality according to the theory of the \bfi{}. 
Some experimental setups lead to zero variance of some item responses over all $n=100$ prompts. Therefore, the following items has been dropped for the visualization of the PCA and for the CFA.
\begin{itemize}
    \item A37(-1) - \chatgpt{} personas in-context
    \item A7,E56 - \chatgpt{} in-context seeded
    \item A7 - \gptfour{} in-context seeded
\end{itemize}
Even though this alters the structural models a little, it made it possible to compare the results.
\section{Details CFA}\label{si:details-cfa}
For the CFA we used the \textit{lavaan} package in R\footnote{https://lavaan.ugent.be/}. We used the following sturctural equation models for our analysis (examples for extraversion):
\paragraph{Single-Component}
\begin{verbatim}
extraversion <- 'E =~ E0 + E1 + E2 + E3 + E4
    + E5 + E6 + E7 + E8 + E9 + E10 + E11'
\end{verbatim}
Defining a single extraversion component on which all 12 extraversion items load.
\paragraph{3 Sub-Components}
\begin{verbatim}
extraversion <- 'Sociability =~ E0 + E1 + E2 + E3 
                Assertiveness =~ E4 + E6 + E7 + E5 
                EnergyLevel =~ E8 + E9 + E10 + E11
                Sociability ~~ 0*Assertiveness
                Sociability ~~ 0*EnergyLevel
                Assertiveness ~~0*EnergyLevel'
\end{verbatim}
Defining three uncorrelated sub-components with 4 items each.
\paragraph{3 Sub-Components with 1 acquiescence factor}
\begin{verbatim}
extraversion <- 'Sociability =~ E0 + E1 + E2 + E3 
                Assertiveness =~ E4 + E5 + E6 + E7 
                EnergyLevel =~ E8 + E9 + E10 + E11
                general_factor =~ E0 + E1 + E2 + E3 
                + E4 + E5 + E6 + E7 + E8 + E9 + E10 
                + E11
                Sociability ~~ 0*Assertiveness
                Sociability ~~ 0*EnergyLevel
                Assertiveness ~~0*EnergyLevel
                Sociability ~~ 0*general_factor
                Assertiveness ~~ 0*general_factor
                EnergyLevel ~~ 0*general_factor
\end{verbatim}
Defining 3 uncorrelated sub-components with 4 items each and a general factor (acquiescence) which accounts for participants' tendency to agree/disagree. This is the model of the \bfi{} for which fit indices reach acceptable levels. However, in our experiments with LLMs, lavaan did not find a solution.
\paragraph{Full Model}
\begin{verbatim}
full_model <- '
              E =~ E0 + E1 + E2 + E3 + E4 + E5
              + E6 + E7 + E8 + E9 + E10 + E11
              A =~ A0 + A1 + A2 + A3 + A4 + A5
              + A6 + A7 + A8 + A9 + A10 + A11
              C =~ C0 + C1 + C2 + C3 + C4 + C5
              + C6 + C7 + C8 + C9 + C10 + C11
              N =~ N0 + N1 + N2 + N3 + N4 + N5
              + N6 + N7 + N8 + N9 + N10 + N11
              O =~ O0 + O1 + O2 + O3 + O4 + O5
              + O6 + O7 + O8 + O9 + O10 + O11'
\end{verbatim}
This model aims at quantifying the results of our PCA. Defining 5 factor. However, in our PCA we used uncorrelated components (varimax rotated). The CFA with uncorrelated factors of the full models did not converge for the LLM data which is why we used this simpler model.\\
We uploaded all our R scripts and data to our github repository \textbf{obfuscated for double blind peer review. code and data has bin submitted as .zip file}
\section{Results CFA seeded}\label{si:results-seeded} Table \ref{si-tbl:seeded-cfa} shows the fit indices and reliability scores of the seeded experiment. The NA values are a result of nonconvergence of the CFA.
\begin{table*}[ht]
\begin{center}
\scalebox{0.9}{
\centering
\begin{tabular}{ccclllllllll}
\multicolumn{1}{l}{} & \multicolumn{2}{c}{} & \multicolumn{3}{c}{\textbf{Single Component}} & \multicolumn{3}{c}{\textbf{3 Sub-Components}} & \multicolumn{3}{c}{\textbf{Full Model}} \\ \hline
\textbf{Model} & \multicolumn{1}{c}{$\alpha$} & \multicolumn{1}{c}{$\omega_h$} & \multicolumn{1}{c}{CFI} & \multicolumn{1}{c}{TLI} & \multicolumn{1}{c}{RMSEA} & \multicolumn{1}{c}{CFI} & \multicolumn{1}{c}{TLI} & \multicolumn{1}{c}{RMSEA} & \multicolumn{1}{c}{CFI} & \multicolumn{1}{c}{TLI} & \multicolumn{1}{c}{RMSEA} \\ \hline
\llamatwo & \multicolumn{1}{c}{\textbf{0.96}} & \textbf{0.95} & 0.39 & 0.25 & 0.15 & NA & NA & NA & NA & NA & NA \\ \hline
\chatgpt & \multicolumn{1}{c}{0.67} & 0.69 & 0.74 & 0.68 & 0.10 & NA & NA & NA & 0.35 & 0.32 & 0.10 \\ \hline
\gptfour & \multicolumn{1}{c}{0.65} & 0.64 & 0.63 & 0.55 & 0.08 & NA & NA & NA & 0.33 & 0.30 & 0.10 \\ \hline
\end{tabular}}
\caption{Reliability scores ($\alpha$, $\omega_h$) and fit indices (CFI, TLI, RMSEA) per LLM (empty personas and seeded first answer). Values on the left side of the forward slash are in-context results, on the right side are no-context results. Values are averaged over all personality traits for the Single Component and 3 Component model. Acceptable scores are bolded. NA values could not be calculated due to non-convergence or invalid solutions. Note that all reliability scores are not meaningful due to poor model fit, even though the scores are in an acceptable range.}\label{si-tbl:seeded-cfa}
\end{center}
\end{table*}

\newpage

\end{document}